\documentclass[letterpaper]{article} 
\usepackage{aaai2026}  
\usepackage{times}  
\usepackage{helvet}  
\usepackage{courier}  
\usepackage[hyphens]{url}  
\usepackage{graphicx} 
\usepackage{natbib}  
\usepackage{caption} 
\usepackage{algorithm}
\usepackage{algorithmic}
\usepackage{xspace}
\usepackage{subfigure}
\usepackage{enumitem}
\usepackage{booktabs}
\usepackage{xcolor}
\usepackage{multirow}
\usepackage{color}
\usepackage{lipsum}
\usepackage{caption}
\usepackage{bm}
\usepackage{amsmath}
\usepackage{amsthm}
\usepackage{amsfonts}
\usepackage{amssymb}

\usepackage{newfloat}
\usepackage{listings}
\DeclareCaptionStyle{ruled}{labelfont=normalfont,labelsep=colon,strut=off} 
\floatstyle{ruled}
\newfloat{listing}{tb}{lst}{}
\floatname{listing}{Listing}
\title{From Subtle to Significant: Prompt-Driven Self-Improving \\ Optimization in Test-Time Graph OOD Detection}
\author{
    Luzhi Wang\textsuperscript{\rm 1},
    Xuanshuo Fu\textsuperscript{\rm 2},
    He Zhang\textsuperscript{\rm 3},
    Chuang Liu\textsuperscript{\rm 1},
    Xiaobao Wang\textsuperscript{\rm 4},
    Hongbo Liu\textsuperscript{\rm 1}\thanks{Corresponding Author.}
}
\affiliations{
    \textsuperscript{\rm 1}College of Artificial Intelligence, Dalian Maritime University\\
    \textsuperscript{\rm 2}Computer Vision Center, Universitat Autònoma de Barcelona\\
    \textsuperscript{\rm 3}School of Computing Technologies, RMIT University \\
    \textsuperscript{\rm 4}College of Intelligence and Computing, Tianjin University\\
    
    wangluzhi0@gmail.com, xuanshuo@cvc.uab.es, he.zhang@rmit.edu.au, \\ chuangliu@whu.edu.cn, wangxiaobao@tju.edu.cn, lhb@dlmu.edu.cn
}

\usepackage{bibentry}

\begin{document}

\maketitle

\begin{abstract}
Graph Out-of-Distribution (OOD) detection aims to identify whether a test graph deviates from the distribution of graphs observed during training, which is critical for ensuring the reliability of Graph Neural Networks (GNNs) when deployed in open-world scenarios. 
Recent advances in graph OOD detection have focused on test-time training techniques that facilitate OOD detection without accessing potential supervisory information (e.g., training data). 
However, most of these methods employ a one-pass inference paradigm, which prevents them from progressively correcting erroneous predictions to amplify OOD signals.
To this end, we propose a \textbf{S}elf-\textbf{I}mproving \textbf{G}raph \textbf{O}ut-\textbf{o}f-\textbf{D}istribution detector (SIGOOD), which is an unsupervised framework that integrates continuous self-learning with test-time training for effective graph OOD detection. 
Specifically, SIGOOD generates a prompt to construct a prompt-enhanced graph that amplifies potential OOD signals. To optimize prompts, SIGOOD introduces an Energy Preference Optimization (EPO) loss, which leverages energy variations between the original test graph and the prompt-enhanced graph. 
By iteratively optimizing the prompt by involving it into the detection model in a self-improving loop, the resulting optimal prompt-enhanced graph is ultimately used for OOD detection. 
Comprehensive evaluations on 21 real-world datasets confirm the effectiveness and outperformance of our SIGOOD method.
\end{abstract}

\section{Introduction}
Graph neural networks (GNNs) offer a powerful paradigm for graph representation learning \cite{wang2025adagcl,FuDHL0C25,0012WYY0Y25}, which is widely applied in tasks such as sarcasm detection \cite{wang2025elevating,wang2023augmenting}, recommender systems \cite{jin2023dual}, and fraud detection \cite{pan2025label}. Most GNNs assume that training and test graphs are from the same distribution (in-distribution, ID) \cite{ZhangWYPTP24}. However, when these well-trained GNNs are deployed in open-world scenarios, they inevitably encounter out-of-distribution (OOD) graphs \cite{shen2024optimizing}, leading to misprediction risks. For instance, a GNN might misclassify a structurally similar yet distributionally distinct five-membered lactone molecule as aspirin (a typical drug molecule), resulting in prediction errors in drug identification \cite{DBLP:conf/iclr/000400L0D0P025}. Thus, detecting OOD graphs is essential for ensuring the reliability of GNNs.

\begin{figure}[!t]
    \centering
\includegraphics[width=0.85\columnwidth]{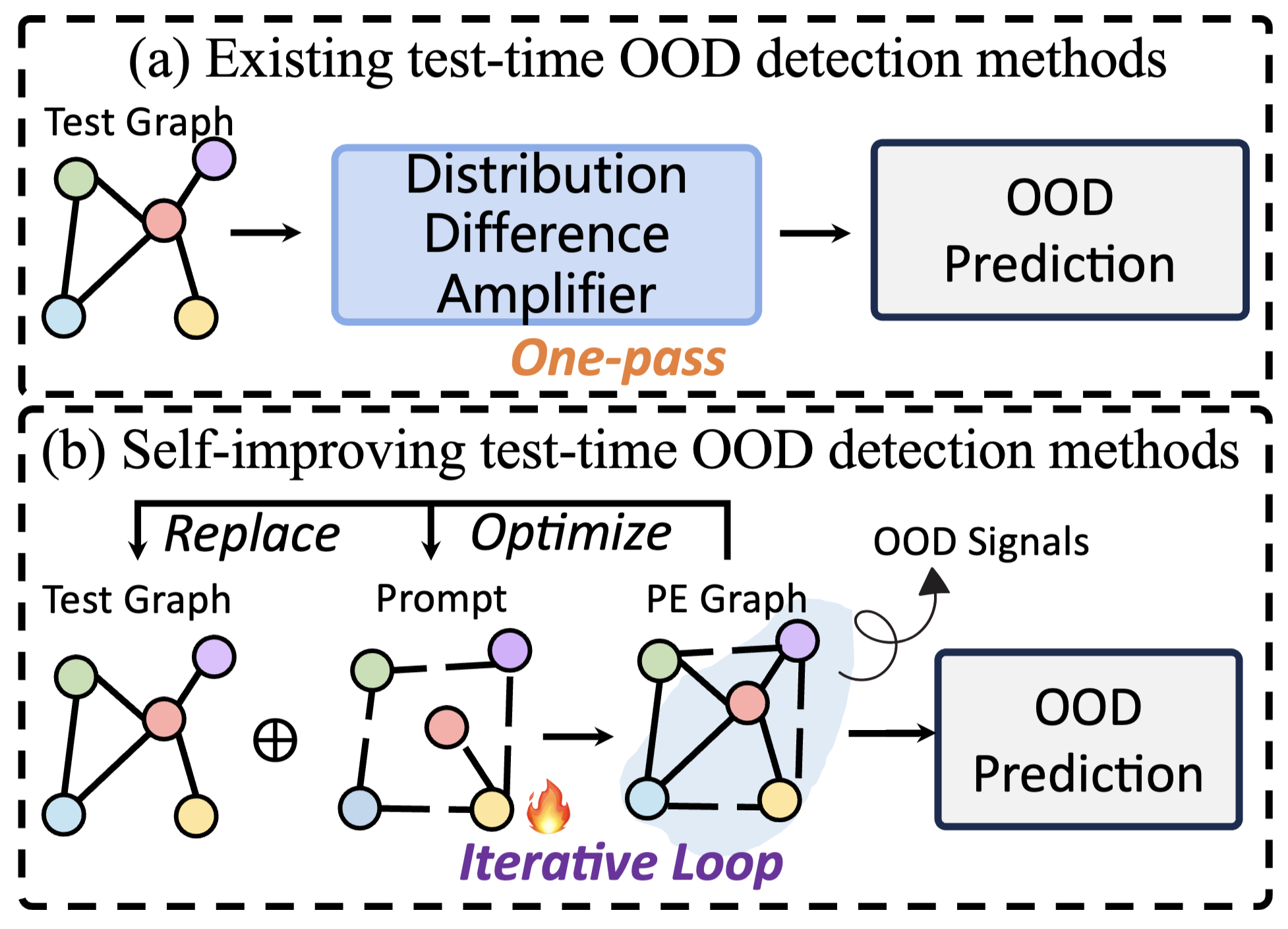}\caption{Comparisons between existing methods and SIGOOD. (a) Existing methods perform a one-pass pattern extraction, aiming to amplify the difference between ID and OOD graphs \cite{wang2024goodat}. (b) SIGOOD adopts an unsupervised self-improvement strategy, progressively refining OOD signals through iterative optimization.
}
    \label{sigood}
\end{figure}
Recent advances in graph OOD detection have attracted increasing attention \cite{lin2025conformal}. 
One line of these studies considers that pattern discrepancies exist between ID and OOD graphs \cite{ding2024sgood}. 
Consequently, these methods rely on training data to learn ID graph patterns for OOD discrimination \cite{hou2025structural}, but such data may be unavailable due to privacy, storage, or deployment constraints \cite{wang2024goodat}.
Therefore, the second line of studies has focused on test-time graph OOD detection, which aims to identify distribution shifts using only the well-trained GNN and test graphs \cite{gtrans}. 
However, these methods are suboptimal when amplifying the difference of ID and OOD core information across graphs, as their one-pass unsupervised rationales neglect to leverage the latent information in test graphs through iterative refinement.


The above limitations motivate us to develop a new test-time graph OOD detection method, but it is not trivial due to the following challenges. 
Particularly, \textit{Challenge 1:} Distribution overlapping. A well-trained GNN primarily learns the decision boundaries from ID graphs, without explicitly acquiring the ability to recognize unknown patterns. Thus, in the embedding space, the GNN tends to project OOD graph embeddings blindly toward the ID graph embeddings, leading to the distributional overlap between ID and OOD data \cite{wu2024graph}. 
This phenomenon makes it difficult to identify OOD graphs that have subtle structural or feature differences from ID graphs. 
\textit{Challenge 2:} Unknown ID/OOD labels. Under the test-time OOD detection setting, the detection model has no access to task labels or other information (e.g., training data), which raises difficulty in effectively extracting OOD patterns with unlabeled test graphs. 
\textit{Challenge 3:} Effective optimization strategies. 
Due to the absence of task labels, it is challenging to design a self-improving optimization strategy that can continuously learn OOD signals from the test graph itself, and effectively enhance the model's OOD detection performance.

To address aforementioned challenges, we propose \textbf{SIGOOD}, an unsupervised self-improving graph OOD detection framework that iteratively enhances OOD signals at test time via energy-based feedback. 
According to previous studies \cite{zhang2024egonc, DBLP:conf/iclr/WuCYY23}, the energy based on prediction logits represents the probability of a graph being OOD, where a high energy score indicates a high likelihood of OOD.
Since OOD signals are composed of key nodes that sufficiently represent OOD characteristics, we introduce prompts to amplify OOD signals by capturing the average node-level energy variations before and after prompt enhancement, thus optimizing the objective of OOD detection. 
Specifically, to address \textit{Challenge 1}, SIGOOD firstly utilizes the well-trained GNN to obtain the embedding of a test graph, and then injects auxiliary prompt into it, aiming to amplify the differences between OOD patterns and ID patterns in it. 
To tackle \textit{Challenge 2}, SIGOOD utilizes the energy variation between the test graph and the prompt-enhanced (PE) graph as a signal to indicate OOD or ID preference, which actively discover and iteratively refine ID and OOD patterns.
To address \textit{Challenge 3}, we propose a tailored loss function that leverages the energy variations induced by prompt injection. This loss amplifies the energy differences between OOD and ID patterns within the PE graph to facilitate more effective OOD detection. 
Unlike prior one-pass methods, our approach involves the prompt-enhanced graph in an iterative optimization loop, supporting a self-improving process.
Fig.~\ref{sigood} illustrates the differences between SIGOOD and other test-time graph OOD detection methods. 
Our contributions are summarized as follows:
\begin{itemize}
\item To the best of our knowledge, we propose the first self-improving framework for test-time graph OOD detection, which progressively enhances the OOD detection performance with only relying on test data.

\item We investigate the role of energy variation in capturing OOD signals and propose an energy preference optimization (EPO) loss to enhance the energy contrast between OOD and ID signals. 
\item Extensive experiments on $21$ benchmark datasets and comparisons with $12$ state-of-the-art methods demonstrate the superior performance of SIGOOD.

\end{itemize}

\section{Related Work}
\subsection{Graph OOD Detection}
Graph OOD detection aims to identify test graphs whose distributions deviate from the ID graphs \cite{lin2024graph,zhang2025conformal}. 
Existing methods generally use pre-trained encoders with post-hoc detectors, such as distance or energy-based scores, which are static detection methods that lack any training-time or test-time optimization process \cite{fuchsgruber2024energy}. 
Several studies focus on improving ID recognition rather than detecting OOD graphs \cite{cao2025ibpl}. Recent test-time methods amplify ID-OOD differences but lack iterative refinement mechanisms to progressively enhance OOD signals \cite{zhang2024fully}. In contrast, our SIGOOD model introduces a novel energy-based self-improving mechanism to iteratively mine OOD patterns without label or training supervision.

\subsection{Test-time Training}
Test-Time Training (TTT) is a general approach designed to enhance the performance of predictive models when there is a distribution shift between training and test data \cite{ liang2025comprehensive}.
In addition to the computer vision \cite{dalal2025one}, natural language processing \cite{zhang2024test}, and multimodal foundation models \cite{bi2025customttt}, TTT technology is also widely used to
improve graph OOD generalization \cite{DBLP:conf/iclr/ZhengSW0P24}. Nevertheless, the application of TTT techniques to graph OOD detection remains largely underexplored.
In our work, we leverage TTT in a self-improving manner to identify OOD signals of graphs, thereby enabling effective test-time OOD detection.
\section{Preliminaries}
\subsubsection{Graphs.}

Given a graph ${G = (V, E, X)}$, the node set is indicated by $V = \{v_1, v_2, \dots, v_n\}$, $E = \{e_1, e_2, \dots, e_m\}$ denotes the set of edges, and $X \in \mathbb{R}^{n \times d}$ represents the $d$-dimensional node feature matrix. Each graph is associated with an adjacency matrix $A \in \mathbb{R}^{n\times n}$, where $A_{ij} = 1$ indicates the presence of an edge between nodes $v_i$ and $v_j$, and $A_{ij} = 0$ indicates that no edge exists between $v_i$ and $v_j$.

\begin{figure*}[!t]
\centering 
\includegraphics[width=0.8\textwidth]{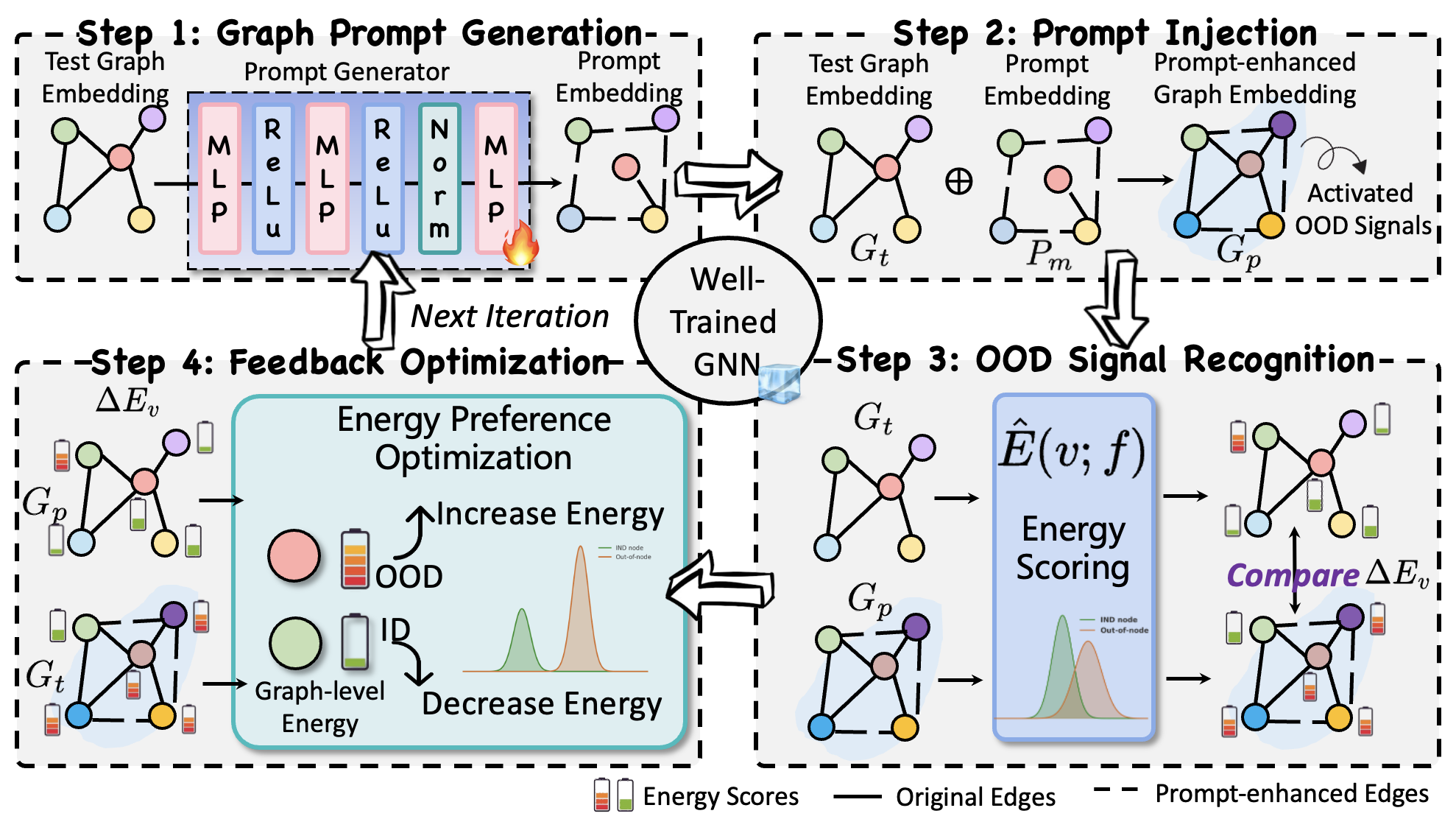}
\caption{Overview of SIGOOD. \textbf{Step 1:} Given a test graph $G_t$ as an input, SIGOOD first encodes it using the well-trained GNN. The obtained embedding is then passed to the prompt generator (PG), which produces a prompt $P_m$ to enhance OOD signals of the graph. 
\textbf{Step 2:} The prompt $P_m$ is applied to $G_t$, yielding a prompt-enhanced (PE) graph $G_p$ with amplified OOD tendency. 
\textbf{Step 3:} Calculate node-wise energy variations between $G_p$ and $G_t$ to locate sensitive nodes of OOD signals.
\textbf{Step 4:} Calculate the global energy variations between $G_p$ and $G_t$ to evaluate the overall OOD tendency of $G_p$. These energy variations are used as OOD signals to guide the optimization of the prompt generator. The updated $G_p$ replaces $G_t$ as the input for the next iteration. After convergence, the final $G_p$ is used to calculate the OOD score for detection.}

\label{framework}
\end{figure*}

\subsubsection{Test-time Graph OOD Detection.}
During the inference stage of a GNN model, the test-time graph out-of-distribution detection task aims to determine whether an input graph is drawn from a distribution different from that of the training data. 
At test time, neither the original training data nor auxiliary information such as task-specific labels for OOD detection is accessible. 
Formally, the test-time graph OOD detection can be defined as:
\begin{equation}
    \texttt{Detection label}=
\begin{cases}
1\ (OOD), & if \ \texttt{D}(G_t) \geqslant \tau  \\ 
0\ (ID), & otherwise \\
\end{cases}
\end{equation}
where $\texttt{D}(G_t)$ denotes a scoring function that measures the discrepancy between the test graph $G_t$ and the in-distribution data, and $\tau$ is a predefined detection threshold.




\section{Method}

\subsection{Overview}
According to existing studies \cite{liu2020energy}, energy scores (e.g., the negative log-likelihood of sample predictions) reflect model confidence during the inference phase. For example, a higher energy value (i.e., a lower likelihood) indicates lower confidence in model predictions, suggesting a greater probability that the test sample is OOD data \cite{chen2023secure}.
In this paper, our intuition is that the subtle difference between ID and OOD data can be amplified by introducing additional prompts, thereby increasing the energy-based distinction for OOD detection.
Due to the inductive bias of a well-trained GNN on ID graphs \cite{wu2020comprehensive}, introducing prompts into a test graph generally leads to a greater energy change in its OOD components (e.g., nodes) than in its ID components. This difference in energy change is then used to iteratively identify potential OOD signals in the test graph and guide the prompt generator toward optimal detection performance.

Fig.~\ref{framework} illustrates the overview of our SIGOOD method. Given a test graph, it is first embedded by a well-trained GNN, and then an optimizable prompt is utilized to generate a prompt-enhanced graph that amplifies the OOD signal. Under the guidance of energy-based loss function (EPO loss), the prompt generator is optimized to enhance OOD signals and the corresponding PE graph is regarded as the updated test graph in the self-improving loop. The specific steps are described in detail below.

\subsection{Step 1: Graph Prompt Generator}
Given a test graph $G_t$, SIGOOD leverages a well-trained GNN to obtain its node embedding $h$. 
However, due to the distribution overlapping between ID and OOD data, it is not trivial to conduct test-time OOD detection by directly using the raw embeddings of test graph.
Therefore, we use a graph prompt generator to dynamically guides and improves the distribution distinguishability of ID and OOD data.
Specifically, the prompt generator $PG(\cdot)$ is implemented as a lightweight three-layer MLP modle \cite{rumelhart1986learning}, which rapidly generates OOD prompts during the testing phase of GNNs. It is formally defined as :
\begin{equation}
\begin{aligned}
&v^* = \texttt{ReLU}(\mathbf{W}_2 \cdot \texttt{ReLU}(\mathbf{W}_1 v + b_1) + b_2),\\
    &PG(v) = \mathbf{W}_3
    \left( \gamma \cdot \frac{ v^*- \mathbf{\mu}}{\sqrt{\mathbf{\sigma}^2 + \epsilon}} + \lambda \right) + b_3,
\end{aligned}
\end{equation}
where $v$ is the node embedding, $\mathbf{W}_1$, $\mathbf{W}_2$, $\mathbf{W}_3$ are trainable weight matrices, and $\gamma$, $\lambda$, $\mu$, $\sigma$ represent learnable parameters for normalization. The prompt $P_m$ is composed of all nodes re-embedded by the prompt generator $PG(v)$.  
These parameters allow the generator to effectively guide the distinguishing on graphs.

\subsection{Step 2: Prompt Injection}
Previous studies have shown that graph components, such as certain nodes or subgraphs, typically contain specific signals that represent their ID or OOD distribution patterns \cite{yu2023mind}. 
Motivated by this, SIGOOD uses the generated prompt to facilitate the recognition of OOD signals included in a test graph $G_t$. 
Specifically, SIGOOD integrates the generated prompt $P_m$ with $G_t$ to construct a prompt-enhanced graph $G_p$:
\begin{equation}
G_p = G_t \oplus P_m,
\end{equation}
where $\oplus$ denotes element-wise addition between the corresponding node embeddings of $G_t$ and its prompt $P_m$. 
By optimizing the prompt generator, $G_p$ is obtained by improving the OOD signals while suppressing the ID signals in $G_t$, thus enhancing the separability between ID and OOD data.

\subsection{Step 3: OOD Signal Recognition}
A well-trained GNN exhibits strong inductive bias toward ID graphs \cite{wu2020comprehensive}. 
Therefore, if a test graph is ID, the representation generated by the well-trained GNN has a high probability to align with the training distribution.
Otherwise, its embeddings are less aligned with the training distribution. 
Given that energy is a transformed form of negative log-probability, existing OOD detection methods often leverage energy as a scoring function to distinguish ID and OOD graphs \cite{jianglearning}. However, directly observing the energy provides only a static information and fails to capture the model's dynamic response after prompt injection. To further exploit energy scores, SIGOOD analyzes node-wise energy variations between the original test graph $G_t$ and the prompt-enhanced graph $G_p$. 

The calculation of energy variations begins by computing the energy scores of each node in both $G_t$ and $G_p$. SIGOOD adds a lightweight scoring head on the well-trained GNN to produce a 2-dimensional logit for representing the ID and OOD preference. 
For a node embedding $v$, let $f(v) \in \mathbb{R}^2$ denote logit function, the energy of node $v$ is defined as:
\begin{equation}
\begin{aligned}
\hat{E}(v) = -\texttt{log} \sum_{i=1}^{2} \texttt{exp}(f_i(v)).
\end{aligned}
\end{equation}

After computing the energy of each node in both the original graph $G_t$ and the PE graph $G_p$, we quantify the energy variation of each corresponding node to emphasize relative changes. Specifically, the energy variation is defined as: 
\begin{equation}
\begin{aligned}
\Delta E_{v} &= \texttt{log}\hat{E}(v; G_p) - \texttt{log}\hat{E}(u; G_t) \\
    &= \texttt{log}\frac{\hat{E}(v;G_p)}{\hat{E}(u;G_t)},
\end{aligned}\label{deta}
\end{equation}
where $u \in G_t$ is the corresponding node of $v \in G_p$. The energy variation $\Delta E_v$ provides a directional indicator for identifying and amplifying potential OOD signals. 
Following the intuition in the overview, SIGOOD identifies nodes with $\Delta E > 0$ as OOD nodes $v_{ood}$, whereas nodes with $\Delta E < 0$ are regarded as ID nodes $v_{id}$. 

\subsection{Step 4: Feedback Optimization}
To optimize SIGOOD, we amplify the energy variation between the prompt-enhanced graph and the original test graph to strengthen the OOD signal. We proposed an energy preference optimization (EPO) loss $\mathcal{L}_{EPO}$, which is defined as:
\begin{equation}
    \begin{aligned}
        \mathcal{L}_{EPO} \ =-\texttt{log} \sigma \Bigg ( \beta \texttt{log}\mathbb{E}_{v_{ood} \sim G_p} \Big[\frac{\hat{E}(v_{ood}; G_p)}{\hat{E}(u_{ood}; G_t)}\Big]  \\
        -\beta \texttt{log}\mathbb{E}_{v_{id} \sim G_p} \Big[\frac{\hat{E}(v_{id}; G_p)}{\hat{E}(u_{id}; G_t)} \Big] \Bigg).
    \end{aligned}
    \label{epo}
\end{equation}
The EPO loss consists with an energy-based Bradley-Terry model \cite{bradley1952rank} and Kullback–Leibler (KL) divergence \cite{kullback1951information}. The implementation of EPO is as follows.

Given the ID/OOD region signal in Step $3$, SIGOOD optimizes the prompt generator to enhance the distinguishability of OOD data. 
An expected loss is that which can increase the difference between ID and OOD signals to facilitate the detection of OOD data. 
An intuitive optimization strategy is the probability $\mathcal{P}$ of OOD signals is more preferred by the model than ID signals. Inspired by the Bradley-Terry model, the probability $\mathcal{P}$ is formulated as
${\mathcal{P}(v_{ood} \succ v_{id}) = \frac{\exp( r(P_m, v_{ood}))}{\texttt{exp}(r(P_m, v_{ood} )) + \texttt{exp}(r(P_m, v_{id}))}}$,
where $r(P_m, v)$ represents a reward function modeling the joint effect of the prompt $P_m$ and node $v$, designed to enhance OOD signals in graphs. Then we maximize the log-likelihood of such pairwise comparisons, the optimization goal is defined as:
\begin{equation}
\texttt{Max}(\texttt{log} \sigma (r(P_m, v_{ood}) -r(P_m, v_{id}))).
\label{loss0}
\end{equation}

SIGOOD aims to optimize the OOD signals in the original graph $G_p$ under the guidance of a reward function $r(\cdot)$, while ensuring that the PE graph $G_p$ does not deviate excessively from $G_t$. 
To achieve this, SIGOOD introduces the Kullback–Leibler (KL) divergence constraint to regularize the difference between $G_p$ and $G_t$. 
Specifically, SIGOOD defines an energy-based KL-divergence as follows:
$D_{KL}=(\hat{E}(v;G_p)||\hat{E}(u;G_t))$. Inspired by DPO \cite{rafailov2023direct} reward mechanisms, the reward function of SIGOOD can be defined as:
\begin{equation}
\begin{aligned}
    r(P_m,v) &= \beta\texttt{log}\frac{\hat{E}(v;G_p)}{\hat{E}(u;G_t)},
\end{aligned}
\label{reward}
\end{equation}
where $u$ is the correspond node of $v$ in graph $G_t$. 
The detailed derivation is provided in the \textit{Appendix}.
Combined with Eq.\eqref{deta}, the reward function $r(P_m,v)$ can be interpreted as the energy variation induced by parameter amplification. This observation suggests that energy variations can serve as an effective optimization objective for enhancing OOD signals.
Furthermore, by incorporating the reward function into Eq.\eqref{loss0}, the optimization goal can be formulated as: 
\begin{equation}
\begin{aligned}
\texttt{Max}\Bigg( \texttt{log} \sigma\bigg(
\beta \texttt{log} \frac{\hat{E}(v_{ood}; G_p)}{\hat{E}(u_{ood}; G_t)}
- \beta \texttt{log} \frac{\hat{E}(v_{id}; G_p)}{\hat{E}(u_{id}; G_t)} \bigg)\Bigg).
\end{aligned}
\end{equation}
To optimize the prompt generator, we contrast the energy variation of OOD and ID nodes between the PE graph $G_p$ and the input graph $G_t$. 
The first term encourages energy of OOD nodes to increase after prompt injection, while the second term penalizes excessive energy variation in ID nodes. By maximizing their difference, SIGOOD promotes OOD-signal responses while maintaining stability for ID nodes. 

While the aforementioned optimization objectives primarily focus on identifying and amplifying local OOD signals, the core task of SIGOOD is graph-level OOD detection. 
Therefore, relying solely on local signals is insufficient. To comprehensively assess whether a test graph deviates from the training distribution, we further incorporate graph-level energy variation as a global optimization signal, aligning with the model’s preference for global energy shifts.
Specifically, we compute the average energy increase and decrease across all nodes in graphs, capturing the overall energy variation to reflect global OOD tendencies. The final EPO loss, is shown in Eq.\eqref{epo}. 
Guided by energy-based feedback, the prompt generator is iteratively updated to amplify the OOD signal in $G_p$. The refined $G_p$ replaces $G_t$ as the new input for subsequent iterations, forming a self-improving loop that continuously strengthens SIGOOD’s ability to detect OOD patterns. After training, the input graph $G_t$ is combined with the optimized prompt to produce a prompt-enhanced graph. This PE graph is then evaluated using the EPO loss, which serves as the OOD detection scores. A predefined threshold $\tau$ is applied to determine whether the graph is ID or OOD.

\subsubsection{Why SIGOOD is effective?}
Well-trained GNNs exhibit strong expressive power in modeling ID patterns, resulting in high confidence when classifying ID graphs. However, they often show low confidence and unstable predictions for OOD graphs. Energy, which serves as a transformed representation of the GNNs output probability, is widely used as an indicator for OOD detection, where higher energy values suggest a greater likelihood of being OOD.
Leveraging this property, SIGOOD initially identifies potential OOD signals through energy variations. Guided by energy-based feedback, the prompt generator is iteratively optimized to produce prompt-enhanced graphs that better expose OOD patterns. These enhanced graphs are then used as new test inputs $G_t$ in the next iteration.
Through this iterative refinement, SIGOOD progressively corrects prior mispredictions, forming a self-improving optimization loop that enhances OOD detection performance over time.

\section{Experiments}
We evaluate the effectiveness of SIGOOD on both graph OOD detection and anomaly detection tasks. The following sections present the datasets, baseline methods, implementation details, and experimental results.

\subsection{Experimental Setups}

\begin{table*}[t]
\centering
\caption{OOD detection results in terms of AUC score ($\%$). The best results are highlighted with \textbf{bold}.} 
\label{oodd}
\resizebox{1\textwidth}{!}{
\begin{tabular}{lccccccccc}
\toprule
ID dataset & BZR & PTC-MR & AIDS & ENZYMES  & Tox21 & FreeSolv  & ClinTox & Esol & \textit{Avg.}  \\
\cmidrule{1-9}
OOD dataset & COX2 & MUTAG & DHFR & PROTEIN  & SIDER & ToxCast  & LIPO & MUV  &\textit{Rank}\\
\midrule
PK-LOF      & $42.22{\scriptstyle\pm8.39}$ & $51.04{\scriptstyle\pm6.04}$ & $50.15{\scriptstyle\pm3.29}$ & $50.47{\scriptstyle\pm2.87}$ & $51.33{\scriptstyle\pm1.81}$ & $49.16{\scriptstyle\pm3.70}$ & $50.00{\scriptstyle\pm2.17}$ & $50.82{\scriptstyle\pm1.48}$ & $11.1$ \\
PK-OCSVM    & $42.55{\scriptstyle\pm8.26}$ & $49.71{\scriptstyle\pm6.58}$ & $50.17{\scriptstyle\pm3.30}$ & $50.46{\scriptstyle\pm2.78}$ & $51.33{\scriptstyle\pm1.81}$ & $48.82{\scriptstyle\pm3.29}$ & $50.06{\scriptstyle\pm2.19}$ & $51.00{\scriptstyle\pm1.33}$ & $11.0$ \\
PK-iF & $51.46{\scriptstyle\pm1.62}$ & $54.29{\scriptstyle\pm4.33}$ & $51.10{\scriptstyle\pm1.43}$ & $51.67{\scriptstyle\pm2.69}$ & $49.87{\scriptstyle\pm0.82}$ & $52.28{\scriptstyle\pm1.87}$ & $50.81{\scriptstyle\pm1.10}$ & $50.85{\scriptstyle\pm3.51}$ & $8.5$ \\
WL-LOF   &   $48.99{\scriptstyle\pm6.20}$ & $53.31{\scriptstyle\pm8.98}$ & $50.77{\scriptstyle\pm2.87}$ & $52.66{\scriptstyle\pm2.47}$ & $51.92{\scriptstyle\pm1.58}$ & $51.47{\scriptstyle\pm4.23}$ & $51.29{\scriptstyle\pm3.40}$ & $51.26{\scriptstyle\pm1.31}$ & $8.0$ \\
WL-OCSVM    & $49.16{\scriptstyle\pm4.51}$ & $53.31{\scriptstyle\pm7.57}$ & $50.98{\scriptstyle\pm2.71}$ & $51.77{\scriptstyle\pm2.21}$ &  $51.08{\scriptstyle\pm1.46}$ & $50.38{\scriptstyle\pm3.81}$ & $50.77{\scriptstyle\pm3.69}$ & $50.97{\scriptstyle\pm1.65}$ & $9.0$ \\
WL-iF  & $50.24{\scriptstyle\pm2.49}$ & $51.43{\scriptstyle\pm2.02}$ & $50.10{\scriptstyle\pm0.44}$ & $51.17{\scriptstyle\pm2.01}$ & $50.25{\scriptstyle\pm0.96}$ & $52.60{\scriptstyle\pm2.38}$ & $50.41{\scriptstyle\pm2.17}$ & $50.61{\scriptstyle\pm1.96}$ & $10.0$ \\
\midrule
InfoGraph-iF & $63.17{\scriptstyle\pm9.74}$ & $51.43{\scriptstyle\pm5.19}$ & $93.10{\scriptstyle\pm1.35}$ & $60.00{\scriptstyle\pm1.83}$ & $56.28{\scriptstyle\pm0.81}$ & $56.92{\scriptstyle\pm1.69}$ & $48.51{\scriptstyle\pm1.87}$ & $54.16{\scriptstyle\pm5.14}$ & $6.4$ \\
InfoGraph-MD & $86.14{\scriptstyle\pm6.77}$ & $50.79{\scriptstyle\pm8.49}$ & $69.02{\scriptstyle\pm11.67}$ & $55.25{\scriptstyle\pm3.51}$ & $59.97{\scriptstyle\pm2.06}$ & $58.05{\scriptstyle\pm5.46}$ & $48.12{\scriptstyle\pm5.72}$ & $77.57{\scriptstyle\pm1.69}$ & $6.1$ \\
GraphCL-iF & $60.00{\scriptstyle\pm3.81}$ & $50.86{\scriptstyle\pm4.30}$ & $92.90{\scriptstyle\pm1.21}$ & $61.33{\scriptstyle\pm2.27}$ & $56.81{\scriptstyle\pm0.97}$ & $55.55{\scriptstyle\pm2.71}$ & $47.84{\scriptstyle\pm0.92}$ & $62.12{\scriptstyle\pm4.01}$ & $7.0$ \\
GraphCL-MD & $83.64{\scriptstyle\pm6.00}$ & $73.03{\scriptstyle\pm2.38}$ & $93.75{\scriptstyle\pm2.13}$ & $52.87{\scriptstyle\pm6.11}$ & $58.30{\scriptstyle\pm1.52}$ & $60.31{\scriptstyle\pm5.24}$ & $51.58{\scriptstyle\pm3.64}$ & $78.73{\scriptstyle\pm1.40}$ & $3.8$ \\
\midrule
GTrans & $55.17{\scriptstyle\pm5.04}$ & $62.38{\scriptstyle\pm2.36}$ & $60.12{\scriptstyle\pm1.98}$ & $49.94{\scriptstyle\pm5.67}$ & $61.67{\scriptstyle\pm0.73}$ & $50.81{\scriptstyle\pm3.03}$ & $58.54{\scriptstyle\pm2.38}$ & $76.31{\scriptstyle\pm3.85}$ & $6.5$ \\

GOODAT & $82.16{\scriptstyle\pm0.15}$ & $81.84{\scriptstyle\pm0.57}$ & $96.43{\scriptstyle\pm0.25}$ & $66.29{\scriptstyle\pm1.54}$ & $68.92{\scriptstyle\pm0.01}$ & $68.83{\scriptstyle\pm0.02}$ & $62.46{\scriptstyle\pm0.54}$ & $85.91{\scriptstyle\pm0.27}$ & $2.3$ \\

\midrule
Ours & $\mathbf{87.36{\scriptstyle\pm0.17}}$ & $\mathbf{85.70{\scriptstyle\pm0.03}}$ & $\mathbf{97.38{\scriptstyle\pm0.01}}$ & $\mathbf{67.88{\scriptstyle\pm0.04}}$ & $\mathbf{69.97{\scriptstyle\pm0.68}}$ & $\mathbf{68.89{\scriptstyle\pm0.24}}$ & $\mathbf{71.33{\scriptstyle\pm0.14}}$ & $\mathbf{87.72{\scriptstyle\pm0.05}}$ & $\mathbf{1}$ \\

\textit{Improve}  &$1.41\%$ & $4.72\%$ & $0.99\%$ & $2.40\%$  & $1.52\%$ & $ 0.09\%$ & $14.20\%$ & $2.11\%$ &$-$  \\
\bottomrule
\end{tabular}}
\end{table*}

\begin{table*}[t!]
\centering
\caption{Anomaly detection results in terms of AUC score ($\%$). The best results are highlighted with \textbf{bold}.} 
\label{ano}
\resizebox{1\textwidth}{!}{
\begin{tabular}{l  cccc cc c c c c}
\toprule
Method & PK-OCSVM & PK-iF & WL-OCSVM & WL-iF & InfoGraph-iF & GraphCL-iF & GTrans & GOODAT & Ours & \textit{Improve}\\
\midrule
PROTEINS-full & $50.49{\scriptstyle\pm4.92}$ & $60.70{\scriptstyle\pm2.55}$ & $51.35{\scriptstyle\pm4.35}$ & $61.36{\scriptstyle\pm2.54}$ & $57.47{\scriptstyle\pm3.03}$ & $60.18{\scriptstyle\pm2.53}$ & $60.16{\scriptstyle\pm5.06}$ & $77.92{\scriptstyle\pm2.37}$ &$79.54 {\scriptstyle\pm0.60}$& $ 2.07\%$ \\
ENZYMES       & $53.67{\scriptstyle\pm2.66}$ & $51.30{\scriptstyle\pm2.01}$ & $55.24{\scriptstyle\pm2.66}$ & $51.60{\scriptstyle\pm3.81}$ & $53.80{\scriptstyle\pm4.50}$ & $53.60{\scriptstyle\pm4.88}$ & $38.02{\scriptstyle\pm6.24}$ & $52.33{\scriptstyle\pm4.74}$ & $7\mathbf{6.80{\scriptstyle\pm0.89}}$ & $39.02\%$\\
DHFR          & $47.91{\scriptstyle\pm3.76}$ & $52.11{\scriptstyle\pm3.96}$ & $50.24{\scriptstyle\pm3.13}$ & $50.29{\scriptstyle\pm2.77}$ & $52.68{\scriptstyle\pm3.21}$ & $51.10{\scriptstyle\pm2.35}$ & $61.15{\scriptstyle\pm2.87}$ & $61.52{\scriptstyle\pm2.86}$ &$\mathbf{65.17{\scriptstyle\pm0.45}}$ & $5.93\%$\\
BZR           & $46.85{\scriptstyle\pm5.31}$ & $55.32{\scriptstyle\pm6.18}$ & $50.56{\scriptstyle\pm5.87}$ & $52.46{\scriptstyle\pm3.30}$ & $63.31{\scriptstyle\pm8.52}$ & $60.24{\scriptstyle\pm5.37}$ & $51.97{\scriptstyle\pm8.15}$ & $64.77{\scriptstyle\pm3.87}$ &$\mathbf{75.42{\scriptstyle\pm3.87}}$ & $16.44\%$\\
COX2          & $50.27{\scriptstyle\pm7.91}$ & $50.05{\scriptstyle\pm2.06}$ & $49.86{\scriptstyle\pm7.43}$ & $50.27{\scriptstyle\pm0.34}$ & $53.36{\scriptstyle\pm8.86}$ & $52.01{\scriptstyle\pm3.17}$ & $53.56{\scriptstyle\pm3.47}$ & $59.99{\scriptstyle\pm9.76}$ &$\mathbf{77.78{\scriptstyle\pm1.85}}$ &$29.65\%$\\
DD            & $48.30{\scriptstyle\pm3.98}$ & $71.32{\scriptstyle\pm2.41}$ & $47.99{\scriptstyle\pm4.09}$ & $70.31{\scriptstyle\pm1.09}$ & $55.80{\scriptstyle\pm1.77}$ & $59.32{\scriptstyle\pm3.92}$ & $76.73{\scriptstyle\pm2.83}$ & $\mathbf{77.62{\scriptstyle\pm2.88}}$ &$72.59{\scriptstyle\pm1.84}$& $-$\\
NCI1          & $49.90{\scriptstyle\pm1.18}$ & $50.58{\scriptstyle\pm1.38}$ & $50.63{\scriptstyle\pm1.22}$ & $50.74{\scriptstyle\pm1.70}$ & $50.10{\scriptstyle\pm0.87}$ & $49.88{\scriptstyle\pm0.53}$ & $41.42{\scriptstyle\pm2.16}$ & $45.96{\scriptstyle\pm2.42}$  &$\mathbf{59.07{\scriptstyle\pm0.47}}$  &$16.41\%$\\
IMDB-B        & $50.75{\scriptstyle\pm3.10}$ & $50.80{\scriptstyle\pm3.17}$ & $54.08{\scriptstyle\pm5.19}$ & $50.20{\scriptstyle\pm0.40}$ & $56.50{\scriptstyle\pm3.58}$ & $56.50{\scriptstyle\pm4.90}$ & $45.34{\scriptstyle\pm3.75}$ & $65.46{\scriptstyle\pm4.34}$ &$\mathbf{68.96{\scriptstyle\pm0.05}}$  & $5.07\%$\\
REDDIT-B      & $45.68{\scriptstyle\pm2.24}$ & $46.72{\scriptstyle\pm3.42}$ & $49.31{\scriptstyle\pm2.33}$ & $48.26{\scriptstyle\pm0.32}$ & $68.50{\scriptstyle\pm5.56}$ & $71.80{\scriptstyle\pm4.38}$ & $69.71{\scriptstyle\pm2.21}$ & $80.31{\scriptstyle\pm0.85}$ &$\mathbf{86.64{\scriptstyle\pm1.97}}$ &$7.88\%$\\
HSE           & $57.02{\scriptstyle\pm8.42}$ & $56.87{\scriptstyle\pm10.51}$ & $62.72{\scriptstyle\pm10.13}$ & $53.02{\scriptstyle\pm5.12}$ & $53.56{\scriptstyle\pm3.98}$ & $51.18{\scriptstyle\pm2.71}$ & $58.49{\scriptstyle\pm2.68}$ & $63.05{\scriptstyle\pm0.90}$ &$\mathbf{64.68{\scriptstyle\pm0.72}}$ &$2.58\%$\\
MMP           & $46.65{\scriptstyle\pm6.31}$ & $50.06{\scriptstyle\pm3.73}$ & $55.24{\scriptstyle\pm3.26}$ & $52.68{\scriptstyle\pm3.34}$ & $54.59{\scriptstyle\pm2.01}$ & $54.54{\scriptstyle\pm1.86}$ & $48.19{\scriptstyle\pm3.74}$ & $69.41{\scriptstyle\pm0.04}$ &$\mathbf{70.17{\scriptstyle\pm0.11}}$ & $1.09\%$\\
p53           & $46.74{\scriptstyle\pm4.88}$ & $50.69{\scriptstyle\pm2.02}$ & $54.59{\scriptstyle\pm4.46}$ & $50.85{\scriptstyle\pm2.16}$ & $52.66{\scriptstyle\pm1.95}$ & $53.29{\scriptstyle\pm2.32}$ & $53.74{\scriptstyle\pm2.98}$ & $\mathbf{63.27{\scriptstyle\pm0.04}}$ &$60.51{\scriptstyle\pm1.95}$ & $-$\\
PPAR-gamma    & $53.94{\scriptstyle\pm6.94}$ & $45.51{\scriptstyle\pm2.58}$ & $57.91{\scriptstyle\pm6.13}$ & $49.60{\scriptstyle\pm0.22}$ & $51.40{\scriptstyle\pm2.53}$ & $50.30{\scriptstyle\pm1.56}$ & $56.20{\scriptstyle\pm1.57}$ & $68.23{\scriptstyle\pm1.54}$ &$\mathbf{72.59{\scriptstyle\pm0.04}}$& $6.39\%$\\
\midrule
\textit{Avg. Rank}    & $7.3$ & $6.3$ & $5.3$ & $6.2$ & $4.9$ & $5.4$ & $5.5$ & $2.6$  &$\mathbf{1.2} $ &$-$ \\
\bottomrule
\end{tabular}}
\end{table*}
\subsubsection{Datasets.}
We selected multiple datasets from diverse domains included in the widely-used UB-GOLD benchmark \cite{DBLP:conf/iclr/000400L0D0P025}. The distribution of OOD detection datasets comes from drug chemical formulas, protein structures, and etc. During testing, ID and OOD samples are mixed in a ${1:1}$ ratio. 
The anomaly detection datasets span biological, chemical, and social domains. Instances belonging to the minority or ground-truth anomaly class are designated as anomalies, whereas others are treated as normal.

\subsubsection{Baselines.}
We compare SIGOOD with $12$ competitive baseline methods, detailed as follows:
\begin{itemize}
    \item \textbf{Traditional OOD Detectors.} These methods employ pre-trained encoders to extract representations, followed by classical OOD or anomaly detectors. Common encoders include the Weisfeiler-Lehman (WL) kernel~\cite{DBLP:journals/jmlr/ShervashidzeSLMB11} and the propagation kernel (PK)~\cite{DBLP:journals/ml/NeumannGBK16}. The downstream OOD detectors include Local Outlier Factor (LOF)~\cite{DBLP:conf/sigmod/BreunigKNS00}, One-Class SVM (OCSVM)~\cite{DBLP:journals/jmlr/ManevitzY01}, Isolation Forest (iF)~\cite{DBLP:conf/icdm/LiuTZ08}, and Mahalanobis Distance-based detector (MD)~\cite{DBLP:conf/iclr/SehwagCM21}.
    
    \item \textbf{GNNs with Post-hoc Detectors.} These methods generate graph representations using GNNs and perform detection in a post-hoc detectors. The GNN encoders include InfoGraph~\cite{DBLP:conf/iclr/SunHV020} and GraphCL~\cite{DBLP:conf/nips/YouCSCWS20}, while the post-hoc detector typically involves Isolation Forest.

    \item \textbf{Test-time Training Methods.} These methods perform OOD detection during test-time without training datasets. Notable methods include GTrans~\cite{gtrans} and GOODAT~\cite{wang2024goodat}.
\end{itemize}

\subsubsection{Implementation Details.}
Following prior studies, we adopt the Area Under the ROC Curve (AUC) as the primary evaluation metric \cite{wang2024goodat}. All experiments are conducted on an NVIDIA RTX 4090 GPU with 24GB of memory. 
Each experiment is repeated five times to ensure stability. For baseline methods, we use the results reported in their original papers, such as GOODAT.

\subsection{Performance on OOD Detection}
Table~\ref{oodd} reports the AUC scores (\%) for graph OOD detection across eight ID/OOD dataset pairs. Overall, SIGOOD consistently achieves the best performance, outperforming $12$ state-of-the-art (SOTA) baselines across all benchmarks. In particular, \textbf{(1)} Compared to traditional OOD detectors, SIGOOD demonstrates a clear advantage over traditional graph OOD methods such as PK-iF and WL-OCSVM. This illustrates the limitations of traditional shallow OOD detectors in capturing complex distribution shifts in graphs.
\textbf{(2)} Compared to GNN-based baselines, SIGOOD exhibits steady improvements, highlighting the effectiveness of test-time enhancement. This indicates that refining the OOD signal at inference time enhances the model’s ability to capture the patterns of OOD graphs.
\textbf{(3)} Compared with other test-time OOD detection methods, SIGOOD maintains robust performance on biochemical datasets. 
On Tox21/ToxCast and ClinTox/LIPO, it achieves AUCs of $69.97\%$ and $71.33\%$, respectively, surpassing the previous best method GOODAT by $1.52\%$ and $14.20\%$. 
These results confirm the effectiveness of SIGOOD, particularly in different domains graph OOD detection scenarios.

\subsection{Performance on Anomaly Detection}
Table~\ref{ano} reports the AUC scores for anomaly detection on $13$ benchmark datasets with $8$ baselines. Overall, SIGOOD consistently delivers the top performance, ranking first on $11$ out of $13$ datasets and achieving the best average rank ($1.2$) among $8$ competing methods. Specifically, \textbf{(1)} SIGOOD significantly outperforms traditional baselines. These results indicate that SIGOOD is also well-suited for graph anomaly detection. \textbf{(2)} SIGOOD also surpasses GNN-based detectors. On BZR dataset, SIGOOD improves over the strongest neural baseline (e.g., GraphCL-iF) by $19.12\%$. It is suggests that, beyond pretraining or contrastive learning, the energy-based self-improvment mechanism enhances the  SIGOOD to distinguish subtle OOD signals that not captured by existing detectors. \textbf{(3)} Compared to SOTA test-time methods, SIGOOD shows notable gains, with improvements of $39.02\%$ on ENZYMES and $29.65\%$ on COX2, demonstrating the effectiveness of self-improvement mechanism across biochemical and social graphs. \textbf{(4)} While SIGOOD does not achieve the top score on DD and p53, these deviations likely stem from dataset-specific variance rather than fundamental limitations. Overall, although SIGOOD fails to achieve the highest score on a few datasets, it achieves highly competitive average rankings across a variety of benchmarks.

\subsection{Ablation Study}
\begin{figure}
	\centering
	\small 
        \subfigure[{Ablation Study on AUC.}]{\includegraphics[width=0.45\columnwidth]{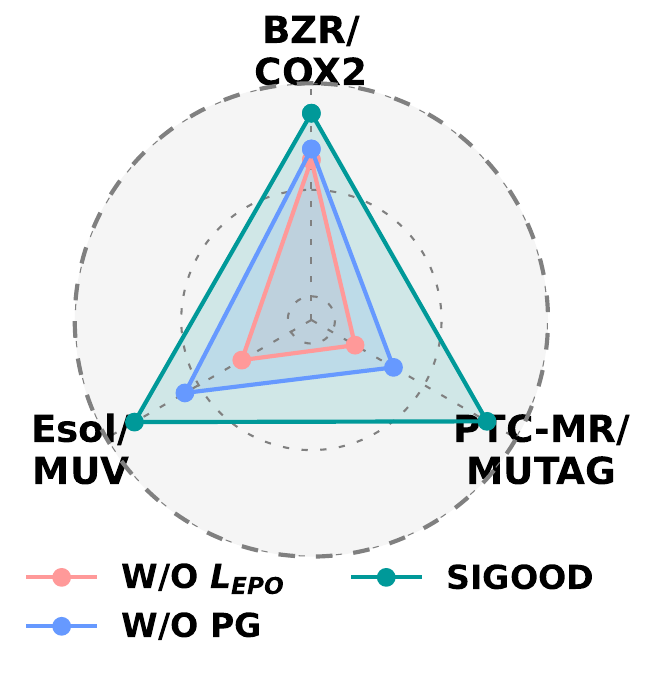}}
	\subfigure[{Effect of PG Depth.}]{\includegraphics[width=0.45\columnwidth]{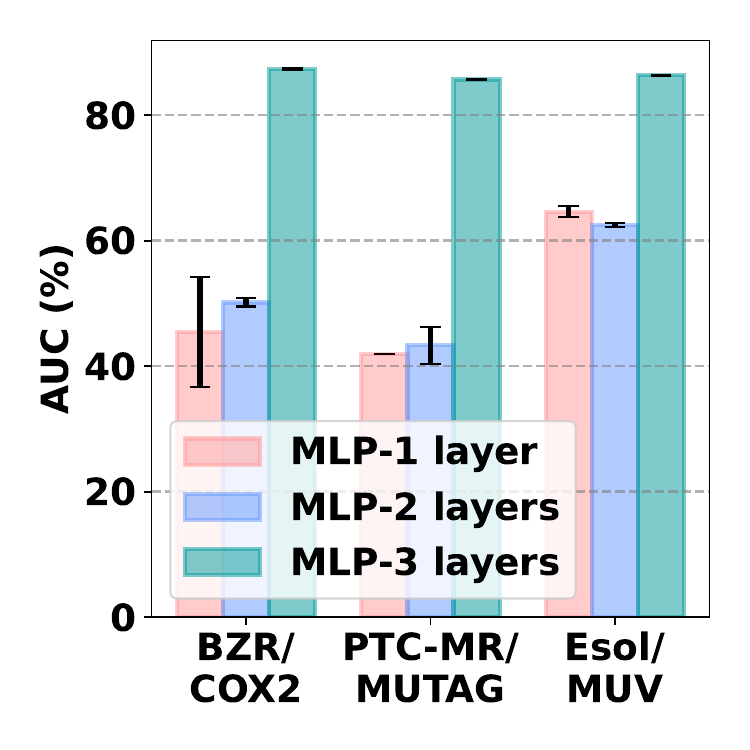}}
	\caption{Ablation Study and PG Depth Analysis.}
	\label{abs}
\end{figure}

The Fig.~\ref{abs} (a) visualizes the performance (AUC) of three model variants across different dataset pairs. As shown, SIGOOD consistently achieves the highest AUC on all three tasks (BZR/COX2, PTC-MR/MUTAG, and Esol/MUV), demonstrating the effectiveness of the full model. In contrast, removing the energy preference optimization loss (W/O $L_{EPO}$) leads to the most severe performance drop, particularly on PTC-MR/MUTAG and Esol/MUV, indicating that $L_{EPO}$ is crucial for enhancing OOD sensitivity. Replacing the prompt generator with an optimizable parameter matrix (W/O PG) also results in a significant performance degradation. This suggests that the prompt generator plays a crucial role in amplifying ood signals.
These results collectively confirm that both the energy-guided optimization and prompt-based modulation are essential to SIGOOD’s superior generalization capability.
\begin{figure}
	\centering
	\small
    	\subfigure[{Effect of Iteration Number.}]{\includegraphics[width=0.46\columnwidth]{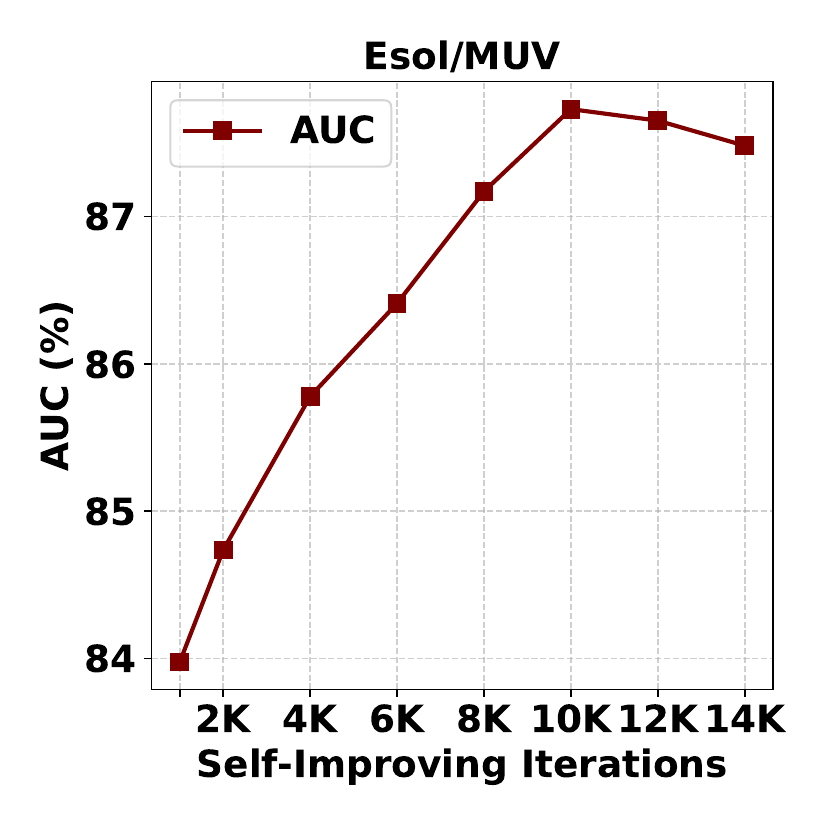}}
        \subfigure[{Effect of the Parameter $\beta$.}]{\includegraphics[width=0.46\columnwidth]{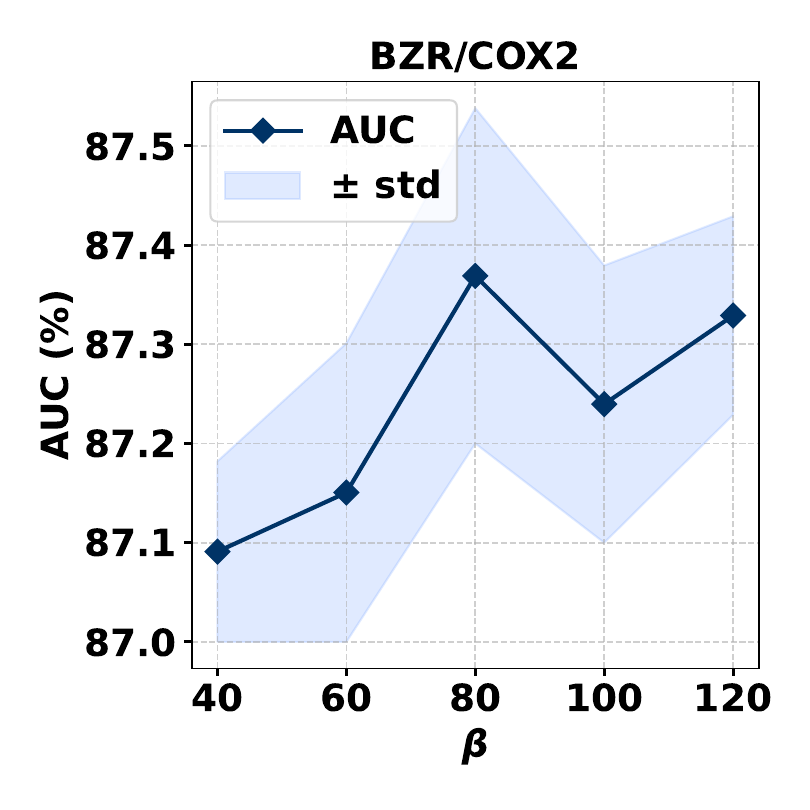}}
	\caption{Parameter Sensitivity Analysis.}
	\label{para}
\end{figure}
\subsection{Parameter Sensitivity Analysis}
\subsubsection{Effect of Prompt Generator (PG) Depth.} To evaluate the impact of the prompt generator depth on OOD detection performance, we experiment with $1$-layer, $2$-layer, and $3$-layer MLPs on three datasets: BZR/COX2, PTC-MR/MUTAG, and Esol/MUV. 
As shown in Fig.~\ref{abs} (b), using a $3$-layer MLP significantly improves detection accuracy, aligning with common empirical practices in deep learning. 
In contrast, shallow configurations yield noticeably lower AUCs, especially on BZR/COX2 and PTC-MR/MUTAG. These results suggest that a deeper prompt generator (with-in $3$-layers) is more effective in capturing complex semantic patterns and refining energy-based representations for reliable OOD detection.

\subsubsection{The effects of Self-improving Interations.}
To intuitively demonstrate the impact of self-improvement iterations on the OOD detection performance, we conducted experiments on the Esol/MUV dataset with large-scale and high-density features. As shown in Fig.~\ref{para} (a), increasing the number of iterations from $1000$ to $10000$ consistently improves the performance of SIGOOD, with the AUC reaching a peak of $87.72\%$. 
This trend demonstrates the effectiveness of iterative on self-learning. However, when the iterations exceed $10000$, the performance gain diminishes slightly, indicating that SIGOOD has approached convergence. Therefore, a moderate iterations offers a favorable trade-off between performance and computational efficiency.

\subsubsection{The effects of parameter $\beta$.}
In SIGOOD, $\beta$ serves as a weighting factor in the energy preference optimization (EPO) objective, controlling the trade-off between preserving original graph semantics and encouraging energy-based differentiation between in-distribution and OOD samples. We investigate the sensitivity of our model to the hyperparameter $\beta$ on the BZR/COX2 dataset. As shown in Fig.~\ref{para} (b), the highest value of $87.36\%$ is observed at $\beta = 80$. This indicates that moderate values of $\beta$ yield slightly better performance. A small $\beta$ may under-emphasize the energy gap, resulting in weaker OOD separation, while a large $\beta$ may distort the embedding space. 
\begin{figure}
    \centering
\includegraphics[width=0.7\columnwidth]{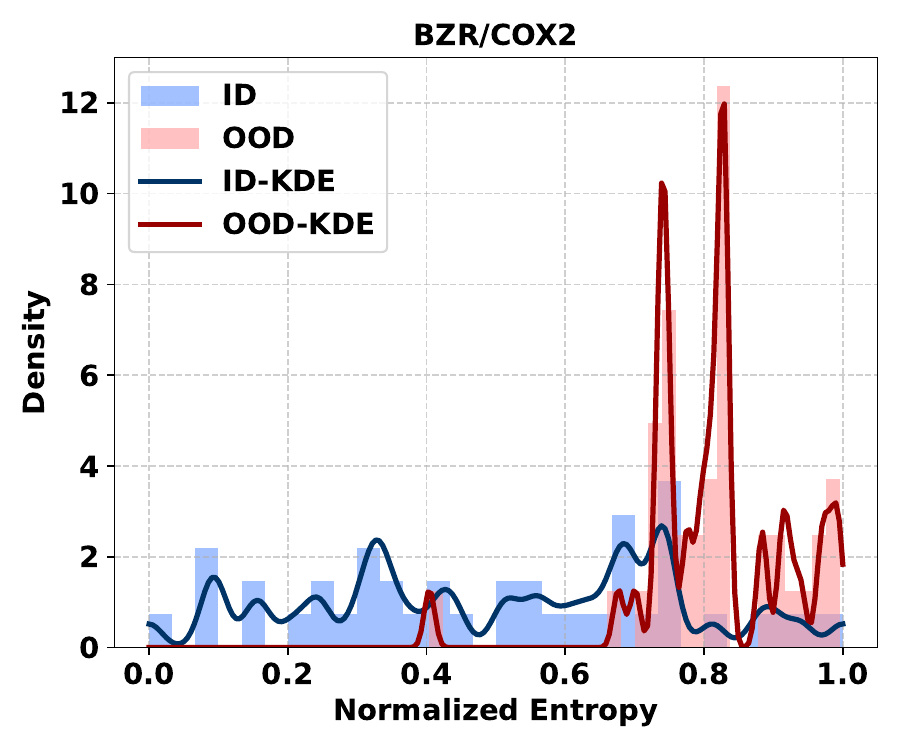}
    \caption{Visualization of Graph Distribution.}
    \label{vis1}
\end{figure}

\subsection{Visualization}
To further illustrate the effectiveness of our method in distinguishing OOD samples, we visualize the normalized entropy distribution of the BZR/COX2 dataset in Fig.~\ref{vis1}. The figure clearly shows a separation between ID and OOD samples, with OOD instances exhibiting higher entropy values. The kernel density estimation (KDE) curves further reveal that the OOD distribution is skewed towards the high-entropy region, while the ID distribution remains concentrated in the low-entropy area. The distinct distributional shift highlights the reliability of entropy as a scoring metric for OOD detection in our framework.

\section{Conclusion}
In this paper, we propose SIGOOD, a self-improving framework for test-time graph OOD detection, which leverages energy-based feedback to iteratively refine OOD detection. 
By integrating a lightweight prompt generator with a well-trained GNN, SIGOOD constructs PE graphs that amplify potential OOD signals. Through energy variations between both the PE graph and the original test graph, the model identifies and enhances OOD-relevant signals. 
SIGOOD also proposes a novel energy preference optimization loss to guide prompt updates, enabling the framework to form a closed-loop self-improving process without requiring additional labels at test time. 
Extensive experiments demonstrate that SIGOOD achieves superior OOD detection performance across various graph datasets, highlighting its effectiveness in real-world scenarios.

\section{Acknowledgments}
This work is supported by National Natural Science Foundation of China (62502065, 62176036, 62302333); the Beatriu de Pinós del Departament de Recerca i Universitats de la Generalitat de Catalunya (2022 BP-00256); the predoctoral program AGAUR-FI ajuts (2025 FI-200470) Joan Oró, which is backed by the Secretariat of Universities and Research of the Department of Research and Universities of the Generalitat of Catalonia, as well as the European Social Plus Fund.



\section{Appendix}
\subsection{Derivation of the Energy-based Reward Function}

We begin with the general form of the optimization objective: 
\begin{equation}
    \max_{\hat{E}(v|P_m)} \; \mathbb{E} \left[ r(P_m,v) \right]
    - \beta \cdot D_{\text{KL}} \left( \hat{E}(v|G_p)|| \hat{E}(u|G_t) \right),
\end{equation}
where $\hat{E}(v; G_p)$ and $\hat{E}(u; G_t)$ denote the energy score from the prompt-enhanced and original test graphs, respectively.
Following the standard expectation formulation $\mathbb{E}x = \sum_x p(x) f(x)$, we treat $\hat{E}(v; G_p)$ as the sampling distribution over nodes, and interpret the KL divergence term as an energy-weighted log-ratio:
\begin{equation}
\begin{aligned}
D_{KL} \left( \hat{E}(v; G_p) || \hat{E}(u; G_t) \right)= \sum_{v} \hat{E}(v; G_p) \texttt{log} \frac{ \hat{E}(v; G_p) }{ \hat{E}(u; G_t) }.
\end{aligned}
\end{equation}
Substituting this into the original objective yields:
\begin{equation}
\max \sum_{v} \hat{E}(v; G_p) \cdot \left[
r(P_m,v) - \beta \texttt{log} \frac{ \hat{E}(v; G_p) }{ \hat{E}(u; G_t) }
\right].
\end{equation}
To maximize the objective, the reward for each outcome $v$ should satisfy:
\begin{equation}
r(P_m,v) = \beta \cdot \texttt{log} \frac{ \hat{E}(v; G_p) }{ \hat{E}(u; G_t) }.
\end{equation}
The above provides the derivation of Eq.\eqref{reward}, which assigns higher rewards to samples exhibiting increased energy after prompt enhancement, indicating a stronger likelihood of being OOD.


\bibliography{aaai2026}

\newpage

\end{document}